\documentclass{article}
\usepackage{spconf,amsmath,graphicx}

\usepackage{amsmath}
\usepackage{comment}
\usepackage{amssymb}
\usepackage{svg}
\usepackage{enumitem}

\usepackage{hyperref}
\setlist[itemize]{label=\textbullet}
\usepackage{multirow}
\usepackage[normalem]{ulem}
\useunder{\uline}{\ul}{}
\usepackage{booktabs}
\usepackage{float}

\usepackage[skip=5pt]{caption} % 调整图表标题间距
\usepackage{setspace}
\usepackage{wrapfig}

\title{EPE-P: Evidence-based Parameter-efficient Prompting for Multimodal Learning with Missing Modalities}
\name{Zhe Chen$^{1}$, Xun Lin$^{1}$, Yawen Cui$^{2}$, Zitong Yu$^{1 *}$\thanks{* Corresponding author}} 

\address{
    $^{1}$Great Bay University \ \ 
    $^{2}$The Hong Kong Polytechnic University \ \ 
}

\begin{document}

\maketitle

\begin{abstract}
Missing modalities are a common challenge in real-world multimodal learning scenarios, occurring during both training and testing. Existing methods for managing missing modalities often require the design of separate prompts for each modality or missing case, leading to complex designs and a substantial increase in the number of parameters to be learned. As the number of modalities grows, these methods become increasingly inefficient due to parameter redundancy. To address these issues, we propose Evidence-based Parameter-Efficient Prompting (EPE-P), a novel and parameter-efficient method for pretrained multimodal networks. Our approach introduces a streamlined design that integrates prompting information across different modalities, reducing complexity and mitigating redundant parameters. Furthermore, we propose an Evidence-based Loss function to better handle the uncertainty associated with missing modalities, improving the model’s decision-making. Our experiments demonstrate that EPE-P outperforms existing prompting-based methods in terms of both effectiveness and efficiency. The code is released at \href{https://github.com/Boris-Jobs/EPE-P\_MLLMs-Robustness}{https://github.com/Boris-Jobs/EPE-P\_MLLMs-Robustness}.
\end{abstract}


\begin{figure}[!t]
\centering
%\vspace{-1.3em}
\includegraphics[width=0.45\textwidth]{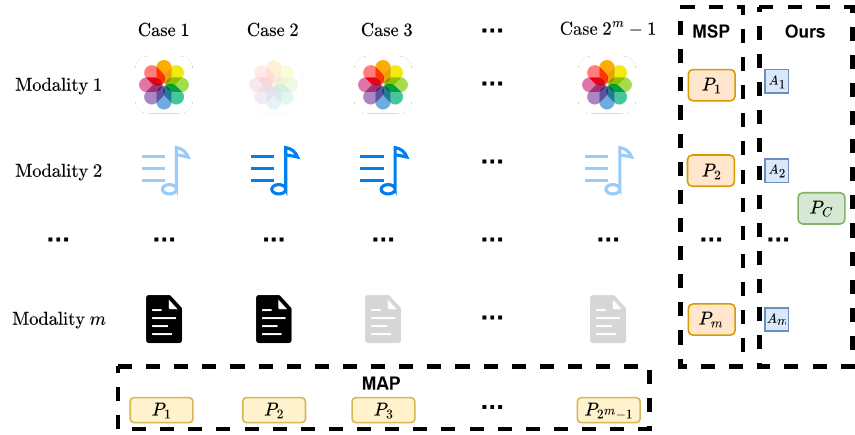}
\caption{The figure illustrates the differences between our approach and existing methods. For a scenario with $m$ modalities, MAP \cite{lee2023multimodalpromptingmissingmodalities} requires designing prompts for each missing case, resulting in a total of $2^m-1$ prompts. MSP \cite{10447257}, on the other hand, requires one prompt per modality, totaling $m$ prompts. Our method only needs a single comprehensive prompt along with $m$ prompt weight matrices, each significantly smaller than a full prompt. This highlights the efficiency of our proposed EPE-P approach.}
\label{model}
\vspace{-1.8em}
\end{figure}

\begin{figure*}[!t]
\centering
\vspace{-1.4em}
\includegraphics[width=0.9\textwidth]{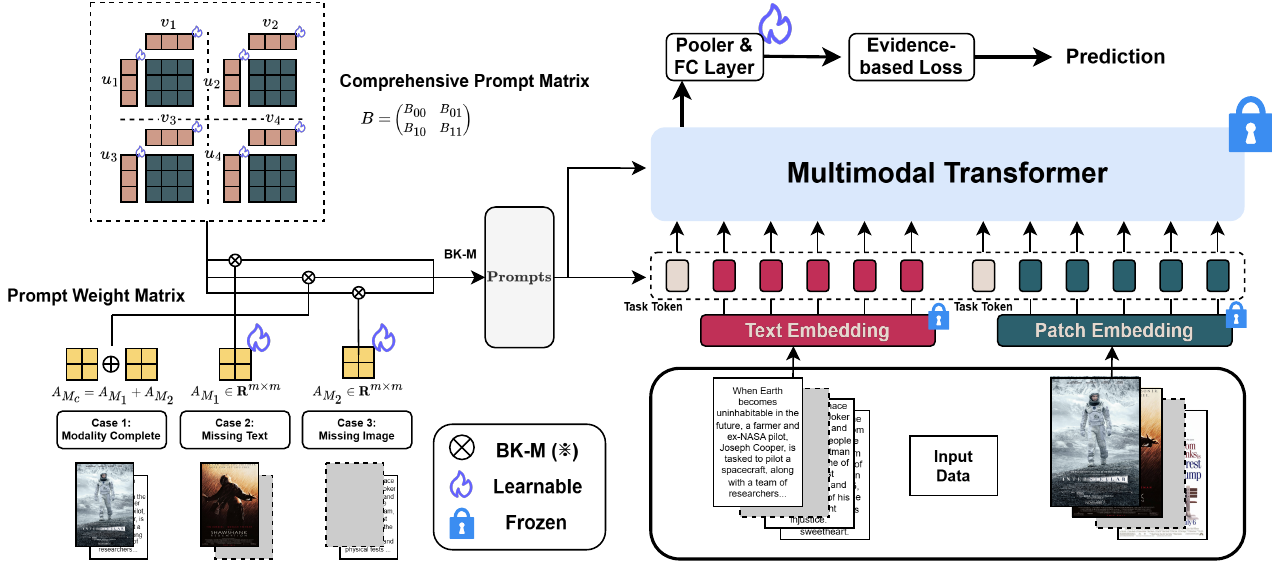}
\caption{Overview of our proposed EPE-P approach for multimodal learning with missing modalities.}
\label{model}
\vspace{-1.4em}
\end{figure*}

\vspace{-0.7em}
\section{Introduction}
\vspace{-0.5em}
\label{introduction}

Multimodal learning utilizes data from various sources to enhance task performance. In real-world applications, input data is often incomplete \cite{ma2021smilmultimodallearningseverely, wu2024multimodal, he2024a, yu2023rethinkingvisiontransformermasked, liu2023fmvitflexiblemodalvision, yu2023flexiblemodalfaceantispoofingbenchmark}, with certain modalities missing. To address this, some methods employ prompt learning \cite{lester2021powerscaleparameterefficientprompt, lee2023multimodalpromptingmissingmodalities}. For instance, MAP \cite{lee2023multimodalpromptingmissingmodalities} faces complexity in designing $2^{m}-1$ prompts for all possible missing cases, where $m$ is the number of modalities. In contrast, MSP \cite{10447257} simplifies the design by creating $m$ prompts and assigning each prompt to a specific modality, but this increases parameter redundancy as the number of modalities \( m \) grows. This redundancy can result in undertrained prompts for infrequently missing modalities, potentially leading to model misguidance during testing with these undertrained prompts.

To overcome these challenges, we propose the Evidence-based Parameter-Efficient Prompting (EPE-P) method. Our approach utilizes a comprehensive prompt and introduces a prompt weight matrix for each missing modality, selectively extracting relevant information from a unified prompt. This reduces complexity and mitigates parameter redundancy. Simultaneously, missing modalities often introduce uncertainty \cite{xu2024reliableconflictivemultiviewlearning}. For example, with a thriller film, a text-only analysis might mistakenly categorize it as a comedy, whereas a comprehensive judgment requires all modalities, such as text and images. Based on Evidential Deep Learning \cite{10.5555/3031657}, we design a loss function that allows both the prompts and the model to integrate uncertainty information for improved decision-making.

Our key contributions are as follows. First, we introduce EPE-P, which utilizes prompt weight matrices to extract information from a comprehensive prompt, reducing design complexity and minimizing parameter redundancy. Second, we design a new loss function based on Evidential Deep Learning theory \cite{Sensoy2018EvidentialDL}, enabling the model and prompts to effectively utilize uncertainty information for better decision-making. Third, we perform extensive experiments on the MM-IMDb \cite{Arevalo2017GatedMU} and Hateful Memes \cite{kiela2020hateful} datasets, demonstrating the effectiveness of our method in managing missing modalities and optimizing prompt usage for complete modality samples during inference.

\vspace{-1.5em}
\section{Methodology}
\vspace{-0.5em}
We design a unique $m \times m$ prompt weight matrix $A_{M_{i}}$ for each missing modality and introduce a comprehensive prompt matrix $B$. Using our Block-wise Kronecker-like Multiplication (BK-M) method, we generate specific prompts tailored to missing cases.

Given a real-world dataset, each sample is represented as $X_i = \{x^{i}_{M_1}, x^{i}_{M_2}, \cdots, x^{i}_{M_m}, y^i\}$, where $m$ is the number of modalities, $x^{i}_{M_j}$ is the $j$-th modality of the $i$-th sample, and $y^i$ is the label. For missing modalities, we use dummy inputs (e.g., empty strings for text, blank pixels for images) and associate each with a prompt weight matrix $A_{M_{i}}$.

For a sample \( X_i \) with missing modalities \( o \) and \( p \), we first sum the matrices \( A_{M_o} \) and \( A_{M_p} \) to obtain \( A_i \). Using the BK-M method, we then compute the BK-M result of \( A_i \) with the comprehensive prompt \( B \) to generate the final prompt. This final prompt, when concatenated with the input embedding information, is used as the input to the multimodal model. For our analysis and experiments, we utilize the multimodal transformer ViLT \cite{kim2021viltvisionandlanguagetransformerconvolution} as the backbone model.

\vspace{-1.2em}
\subsection{Prompt Design}
\vspace{-0.4em}
\indent \textbf{Block-wise Kronecker-like Multiplication (BK-M). }Recent studies have explored the use of the Kronecker Product for compressing model parameters \cite{Edalati2021KroneckerDF, Tahaei2021KroneckerBERTLK}. We propose a novel matrix multiplication operation, termed \textit{Block-wise Kronecker-like Multiplication} (BK-M), denoted by \(\divideontimes\). Given \(A \in \mathbb{R}^{m \times m}\) and \(B \in \mathbb{R}^{d \times l}\), where \(B = \{B_{ij}\}_{m \times m}\) is partitioned into \(m \times m\) blocks, \(B_{ij} \in \mathbb{R}^{\frac{d}{m} \times \frac{l}{m}}\), BK-M is defined as:
 \vspace{-0.2em}
\begin{equation}
\small
    A \divideontimes B = \{a_{ij}B_{ij}\}_{m \times m}
\end{equation}

 \vspace{-0.2em}
\noindent where \(a_{ij}\) denotes elements of \(A\), and \(B_{ij}\) are the submatrices of \(B\). The resulting matrix \(A \divideontimes B\) retains the dimensions of \(B\), i.e., \(d \times l\). Unlike the Kronecker Product, where each element of \(A\) scales the entire matrix \(B\), BK-M scales only the corresponding submatrix \(B_{ij}\).

\noindent \textbf{Local Intrinsic Dimension.} Inspired by \cite{he2023parameterefficientmodeladaptationvision}, which introduced the concept of local intrinsic dimension focusing on the intrinsic dimensions of submodules, we adopt a similar approach. They demonstrated that optimizing models in a low-rank subspace can achieve strong performance. In line with this, we decompose the comprehensive prompt matrix \( B \) as follows:
\vspace{-0.4em}
\begin{equation}
\small
    B_{ij} = u_{ij}v_{ij}^T,
\end{equation}

\vspace{-0.4em}
\noindent where \( u_{ij} \in \mathbb{R}^{\frac{d}{m} \times r} \) and \( v_{ij} \in \mathbb{R}^{\frac{l}{m} \times r} \), with \( r \) representing the local intrinsic dimension of the prompts. This decomposition significantly reduces the number of trainable parameters in \( B \), thereby enhancing parameter efficiency, as discussed in Section \ref{efficiency}.

\noindent \textbf{Prompt Details.} We adopt the strategy proposed by MAP \cite{lee2023multimodalpromptingmissingmodalities}, which suggests that integrating prompts with the input of the first 6 transformer layers yields optimal performance:
\vspace*{-0.7em}
\begin{equation}
\small
    h_{i} = E_{i-1}([\textit{prompts};\ h_{i-1}]), \quad \{i=0, \cdots, 5\},
\end{equation}

\vspace*{-0.7em}
\noindent where \(h_{i}\) denotes the \(i\)-th hidden layer, and \([\textit{prompts};\ h_{i-1}]\) represents the concatenation of the prompts with the \((i-1)\)-th hidden layer. Here, \(h_0\) is the input data. The EPE-P \(\textit{prompts}\) is defined based on BK-M as follows:
\vspace*{-6pt}
\begin{equation}
\small
    \textit{prompts} = \left( \sum_{i=1}^{m} \mathbf{1}_{\{M_{1}, \cdots, M_{m}\}}(M_{i}) A_{M_{i}} \right) \divideontimes B,
\end{equation}

\vspace*{-0.5em}
\noindent where \(\mathbf{1}_{\{M_{1}, \cdots, M_{m}\}}(M_{i})\) is an indicator function defined as:
\vspace*{-6pt}
\begin{equation}
\small
    \mathbf{1}_{\{M_{1}, \cdots, M_{m}\}}(M_{i}) = 
    \begin{cases} 
    1 & \text{if } M_{i} \in \{M_{1}, \cdots, M_{m}\}, \\
    0 & \text{if } M_{i} \notin \{M_{1}, \cdots, M_{m}\}.
    \end{cases}.
\end{equation}

\vspace{-1.8em}
\subsection{Evidence-based Loss Function}
\vspace{-0.5em}

Evidential Deep Learning (EDL) \cite{Sensoy2018EvidentialDL} introduces a framework that integrates predictive uncertainty information to mitigate overconfidence in incorrect classifications, which is often caused by the softmax, thereby enhancing model robustness. Similarly, multimodal models with missing modalities naturally involve substantial uncertainty. Our goal with the EPE-P method is not only to compensate for missing modality information but also to incorporate uncertainty information to improve model decision-making. Hence, we propose the Evidence-based Loss Function, $L_{eb}$.

Consistent with \cite{Sensoy2018EvidentialDL}, we use $ReLU(logits)$ as the evidence $\boldsymbol{e} \in \mathbb{R}_{+}^K$, where $logits \in \mathbb{R}^K$ denotes the output of ViLT \cite{kim2021viltvisionandlanguagetransformerconvolution}, $K$ is the number of classes. Through Type II Maximum Likelihood estimation applied to cross-entropy, the Evidence-based Loss Function $L_{eb}$ can be expressed as follows \cite{Sensoy2018EvidentialDL}:

\vspace{-2em}
\begin{align}
\small
\nonumber \mathit{L_{eb}} &= \int \left [ \sum_{j=1}^{K}-y_{j}\log p_{j} \right ] \frac{\textstyle \prod_{j=1}^{K}p_{j}^{{\alpha}_{j}-1}}{B(\boldsymbol{\alpha})} \, d\boldsymbol{p} \\ 
&= \sum_{j=1}^{K} y_{j} (\psi(S) - \psi(\alpha_{j})),
\end{align}

\vspace{-0.9em}
\noindent where $\boldsymbol{\alpha} = \boldsymbol{e} + 1$ represents the parameters of the Dirichlet distribution, and $\psi(\cdot)$ is the digamma function. $S$ is the sum of all elements in $\boldsymbol{\alpha}$. Here, $\boldsymbol{p} = \frac{\boldsymbol{\alpha}}{S} \in [0, 1]^K$ denotes the class probabilities and $\mathbf{y}$ denotes the label.

To address the issue of incorrect labels potentially generating higher evidence, we introduce an additional Kullback-Leibler (KL) divergence term:
\vspace{-0.7em}
\begin{equation}
L_{KL} = KL \left[D(\boldsymbol{p}|\tilde{\boldsymbol{\alpha}}) \parallel D(\boldsymbol{p}|\mathbf{1})\right],
\end{equation}

\vspace{-0.7em}
\noindent where $D(\boldsymbol{p}|\mathbf{1})$ denotes the uniform Dirichlet distribution, and $\tilde{\boldsymbol{\alpha}} = \mathbf{y} + (\mathbf{1} - \mathbf{y}) \odot \boldsymbol{\alpha}$ adjusts the Dirichlet parameters by accounting for the removal of non-misleading evidence from $\boldsymbol{\alpha}$ for the current sample. This regularization term helps alleviate the adverse effects of misleading evidence \cite{Sensoy2018EvidentialDL}.

\vspace*{-1em}
\subsection{Overall Objective}
\vspace{-5pt}
Therefore, the objective function of our model can be formulated as follows:
\vspace*{-2em}

\begin{align}
\small
    \nonumber \min_{A, u, v, \theta} \quad & L\left(f\left[\left( \sum\limits_{i=1}^{m} \mathbf{1}_{\{M_{1}, \cdots, M_{m}\}}(M_{i}) A_{M_{i}} \right) \divideontimes B; x\right]; \mathbf{y}\right), \\
    \text{s.t.} \quad &  \ B_{ij} = u_{ij}v_{ij}^T, \quad L = (1 - \lambda) L_{eb} + \lambda L_{KL}
\end{align}

\vspace{-5pt}
\noindent where \(A \in \mathbb{R}^{m \times m}\), \(u \in \mathbb{R}^{\frac{d}{m} \times r}\), and \(v \in \mathbb{R}^{\frac{l}{m} \times r}\). Here, \(f\) denotes the baseline model, \(L\) represents the loss function, \(\theta\) includes the parameters of the pooler and fully-connected layers, and \(\mathbf{y}\) is the one-hot encoded label vector.
\vspace*{-10pt}
\subsection{Analysis of Parameter Efficiency}
\label{efficiency}

\begin{table}[]
\centering
\begin{tabular}{ccc}
\hline
Method & \#Params                      & Complexity                \\ \hline
MAP \cite{lee2023multimodalpromptingmissingmodalities}    & $(2^m - 1) \times d \times l$ & $\mathcal{O}(d \times l)$ \\
MSP \cite{10447257}   & $m \times d \times l$         & $\mathcal{O}(d \times l)$ \\
\textbf{EPE-P (Ours)}   & $(d + l) \times r + m^3$      & $\mathcal{O}(d + l)$                   \\ \hline
\end{tabular}
\vspace{-0.4em}
\caption{Efficiency analysis on missing-aware prompts (MAP) \cite{lee2023multimodalpromptingmissingmodalities}, modality-specific prompts (MSP) \cite{10447257}, and our proposed prompting method (EPE-P).}
\label{tab:param-table}
\vspace{-1.4em}
\end{table}
\vspace*{-6pt}

We evaluate the parameter efficiency of our EPE-P method in comparison to two other prompting methods designed for handling missing modalities.

\textbf{Missing-Aware Prompts} \cite{lee2023multimodalpromptingmissingmodalities}: This approach requires designing prompts for all possible missing modality scenarios, resulting in \(2^m - 1\) distinct prompts. Consequently, the total number of parameters for the prompts is \((2^m - 1) \times d \times l\), where \(d\) and \(l\) denote the height and width of each prompt, respectively.

\textbf{Modality-Specific Prompts} \cite{10447257}: This method involves assigning a separate prompt for each modality, leading to a total number of parameters equal to \(m \times d \times l\).

\textbf{EPE-P (Ours)}: In our approach, we utilize a prompt weight matrix \(A_{M_i}\) for each modality, with each matrix containing \(m \times m\) parameters. Additionally, the comprehensive prompt matrix \(B\) has \((d + l) \times r\) parameters. Thus, the total number of parameters in our method is \((d + l) \times r + m^3\).

In real-world applications, given that \(m \ll d, l\) and \(r \ll d, l\), it follows that \((d + l) \times r + m^3 < m \times d \times l < (2^m - 1) \times d \times l\). This demonstrates that our proposed EPE-P method requires fewer parameters and lower computational cost compared to MAP \cite{lee2023multimodalpromptingmissingmodalities} and MSP \cite{10447257}. A detailed summary of these parameter counts is provided in Table \ref{tab:param-table}.

\vspace{-1.0em}
\section{Experiment}
\label{sec:typestyle}
\vspace{-0.6em}

\begin{table*}[t]
\centering
\vspace{-1.2em}
\resizebox{0.86\textwidth}{!}{%
\begin{tabular}{c|c|cc|cc|cccc}
\hline
\multirow{2}{*}{Datasets}                                                           & \multirow{2}{*}{\begin{tabular}[c]{@{}c@{}}Missing-\\ Rate\end{tabular}} & \multicolumn{2}{c|}{Training} & \multicolumn{2}{c|}{Testing} & \multirow{2}{*}{Baseline \cite{kim2021viltvisionandlanguagetransformerconvolution}} & \multirow{2}{*}{MAP \cite{lee2023multimodalpromptingmissingmodalities}} & \multirow{2}{*}{\textbf{EPE-P (Ours)}} & \multirow{2}{*}{EPE-P ($-L_{eb}$)} \\ &                                                                          & Text          & Image         & Text          & Image        &                           & & & \\ \hline
\multirow{6}{*}{\begin{tabular}[c]{@{}c@{}}MM-IMDb\\ (F1-Macro)\end{tabular}}       & \multirow{3}{*}{50\%}                                                    & 100\%         & 50\%          & 100\%         & 50\%         & 40.21                     & \underline{44.67}                & \textbf{46.32} & 43.98 \\ & & 50\%          & 100\%         & 50\%          & 100\%        & 39.24                     & 45.19                & \textbf{46.98}                        & \underline{45.87} \\  & & 75\%          & 75\%          & 75\%          & 75\%         & 42.17                     & 43.89                & \textbf{48.26}                        & \underline{46.75}                     \\ \cline{2-10} & \multirow{3}{*}{60\%}                                                    & 100\%         & 40\%          & 100\%         & 40\%         & 37.25                     & 39.23 & \textbf{42.36} & \underline{40.23} \\ & & 40\%          & 100\%         & 40\%          & 100\%        & 37.11                     & 39.97                & \textbf{43.58} & \underline{40.57} \\ &                                                                          & 70\%          & 70\%          & 70\%          & 70\% & 38.95 & \underline{41.22}                & \textbf{44.17} & 40.86 \\ \hline
\multirow{6}{*}{\begin{tabular}[c]{@{}c@{}}Hateful-\\ Memes\\ (AUROC)\end{tabular}} & \multirow{3}{*}{50\%}                                                    & 100\%         & 50\%          & 100\%         & 50\%         & 63.52                     & 64.79                & \textbf{67.57}                        & \underline{65.81} \\  & & 50\%          & 100\%         & 50\%          & 100\%        & 63.44 & 63.91                & \textbf{67.34} & \underline{66.28} \\ & & 75\%          & 75\%          & 75\%          & 75\%         & 62.76                     & \underline{65.02}  & \textbf{66.11} & 63.49 \\ \cline{2-10}  & \multirow{3}{*}{60\%}                                                    & 100\%         & 40\%          & 100\%         & 40\%         & 59.58                     & 61.33                & \textbf{63.78}  & \underline{62.25}   \\   & & 40\%          & 100\%         & 100\%         & 40\%         & 59.26                     & 62.15                & \textbf{64.13}   & \underline{62.37}    \\ & & 70\%          & 70\%          & 70\%          & 70\%         & 60.14                     & \underline{62.92}     & \textbf{64.94}                    & 61.88                     \\ \hline
\end{tabular}%
}
\vspace{-0.4em}
\caption{Quantitative results of the baseline model (ViLT \cite{kim2021viltvisionandlanguagetransformerconvolution}), missing-aware prompts (MAP \cite{lee2023multimodalpromptingmissingmodalities}), and our proposed method (EPE-P) on the MM-IMDb dataset \cite{Arevalo2017GatedMU} and the Hateful Memes dataset \cite{kiela2020hateful}. \textbf{Bold} values indicate the best performance, while \underline{underlined} values represent the second-best performance. EPE-P ($-L_{eb}$) represents the results of using EPE-P without the Evidence-based Loss.}
\label{tab:comprehensive_table}
\vspace{-1.4em}
\end{table*}

\subsection{Database \& Metrics}
\vspace{-5pt}
In the tables presented in this paper, both F1 Macro and AUROC metrics are expressed as percentages. For example, an F1 Macro score of 40.21\% is displayed as 40.21 in the tables.

\noindent \textbf{MM-IMDb} \cite{Arevalo2017GatedMU} is a multimodal dataset containing text and image modalities for movie genre classification across 23 genres, with multi-label samples. We evaluate performance using the F1-Macro metric. \textbf{Hateful Memes} \cite{kiela2020hateful} is a binary classification dataset comprising text and image modalities, focused on identifying hateful memes. Performance is assessed using AUROC.

\vspace*{-1em}
\subsection{Implementation Details}
\vspace{-5pt}
Our proposed prompting method (EPE-P) is implemented using the PyTorch library. We utilized two NVIDIA A100-SXM4-40GB GPUs for training, with a batch size of 256. The optimization of prompt parameters, as well as the parameters of the pooler and fully connected layers, is performed using the AdamW optimizer \cite{loshchilov2019decoupledweightdecayregularization} with a learning rate of 1e-2, following a cosine annealing schedule. The weight parameter $\lambda$ in Evidence-based Loss $L$ is 0.004. The weight decay is set to 2e-3. For text processing, we employ the `bert-base` tokenizer to tokenize the input text. For image processing, we use the way ViLT processes the images \cite{kim2021viltvisionandlanguagetransformerconvolution}.

\vspace*{-1em}
\subsection{Results and Analysis}
\vspace{-5pt}

\noindent \textbf{Robustness Against Missing Modalities.} We conducted experiments on the MM-IMDb \cite{Arevalo2017GatedMU} and Hateful Memes \cite{kiela2020hateful} datasets with missing rates of 50\% and 60\% as defined in \cite{lee2023multimodalpromptingmissingmodalities}. The results, summarized in Table \ref{tab:comprehensive_table}, show the performance of different methods under the same missing cases for both training and inference. Our proposed method incorporates a comprehensive matrix as the main part of the prompt, which differs from existing methods such as MAP \cite{lee2023multimodalpromptingmissingmodalities} and MSP \cite{10447257}. Unlike these methods, which use different prompts for specific missing cases or modalities, our approach consistently integrates this comprehensive matrix into each input. This allows our prompts to better learn complete information throughout the entire training process. At a 50\% missing rate, our method achieves an average improvement of 2.60 in F1-Macro on MM-IMDb and 2.43 in AUROC on Hateful Memes compared to MAP \cite{lee2023multimodalpromptingmissingmodalities}. At a 60\% missing rate, our approach provides an average improvement of 3.23 in F1-Macro on MM-IMDb and 2.15 in AUROC on Hateful Memes compared to MAP \cite{lee2023multimodalpromptingmissingmodalities}.

\begin{figure}[t]
  \centering
   \vspace{-1.5em}
  \includegraphics[width=0.5\textwidth]{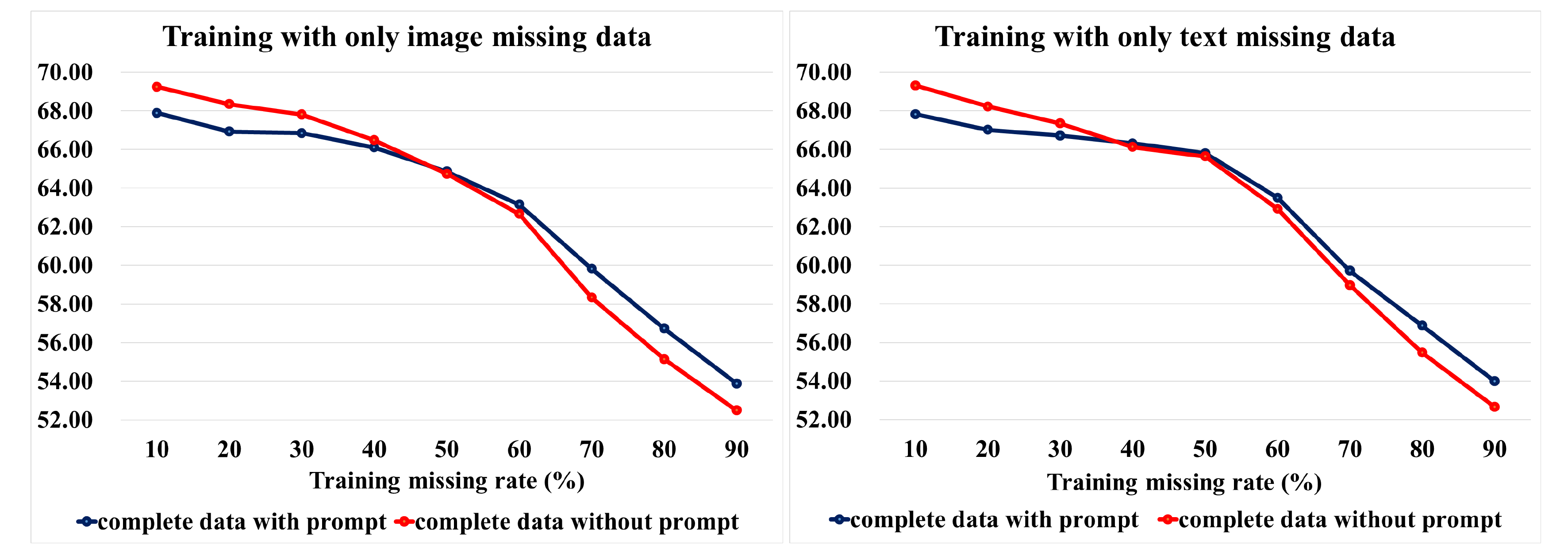}
  \vspace{-1.5em}
  \caption{Quantitative results of proposed EPE-P on the Hateful Memes \cite{kiela2020hateful} dataset with varying missing rates. The evaluation was conducted on a test set with a 25\% missing rate for both text and images.}
  \label{imagelinechart}
\vspace{-1.5em}
\end{figure}

\noindent \textbf{Optimal Use of EPE-P.} As discussed in Section \ref{introduction}, insufficiently trained prompts can mislead inference on complete modality samples. When the missing rate is low, it results in inadequately trained prompts. Therefore, we hypothesized that different prompting strategies might be needed to improve model performance under such conditions. To investigate this, we controlled the training of prompts with varying missing rates. As shown in Figure \ref{imagelinechart}, our findings reveal that when the training missing rate is below 40\%, omitting prompts for complete samples leads to better performance. Conversely, when the training missing rate exceeds 50\%, using prompts for all test samples improves performance. Specifically, when the missing rate is below 30\%, the average AUROC improvements for the two scenarios depicted in Figure \ref{imagelinechart} are 0.89 and 0.76, respectively. Thus, we recommend avoiding the use of prompts for complete samples when the training missing rate is below 30\%, while employing prompts for all test samples is beneficial when the training missing rate is higher than 30\%. This also suggests that inadequately trained prompts can adversely affect inference on complete modality samples and highlights the potential negative impact of parameter redundancy—manifested as an excess of inadequately trained parameters—on overall model performance.

\noindent \textbf{Effectiveness of $L_{eb}$.} In addition to evaluating the performance of the proposed EPE-P and the baseline ViLT \cite{kim2021viltvisionandlanguagetransformerconvolution}, as well as MAP \cite{lee2023multimodalpromptingmissingmodalities}, we conducted experiments to assess the effectiveness of combining EPE-P with Evidence-based Loss. The results are presented in Table \ref{tab:comprehensive_table}. We approached the classification problem as a belief assignment problem, where belief assignments are modeled using a Dirichlet distribution according to Subjective Logic \cite{10.5555/3031657}. This framework allows us to quantify belief masses and uncertainty, thereby guiding the model towards improved classification performance. At a 50\% missing rate, incorporating Evidence-based Loss (EPE-P) results in an average improvement of 1.65 in F1-Macro on MM-IMDb and 1.81 in AUROC on Hateful Memes compared to EPE-P without Evidence-based Loss (EPE-P ($-L_{eb}$)). At a 60\% missing rate, the improvements are 2.82 in F1-Macro on MM-IMDb and 2.12 in AUROC on Hateful Memes. These improvements are consistent across different scenarios, demonstrating enhanced performance with the addition of $L_{eb}$.

\vspace{-1.8em}
\section{Conclusion}
\vspace{-0.9em}

%In this paper, we introduced Evidence-based Parameter-Efficient Prompting (EPE-P), which addresses two primary issues: 1) the complexity of prompt design inherent in traditional methods for managing missing modalities, and 2) the risk of misleading inference due to parameter redundancy caused by excessive prompt parameters. We proposed a streamlined approach that includes a comprehensive prompt matrix integrating information for all potential missing cases and developed specific prompt weight matrices for each modality to selectively extract relevant information from this comprehensive matrix. This approach not only compensates effectively for missing information but also reduces parameter redundancy. Additionally, we introduced the optimal use of EPE-P along with Evidence-based Loss to enhance the model's ability to handle uncertainty and adapt to missing modalities. Extensive experiments demonstrate the effectiveness of our method.

In this paper, we introduced Evidence-based Parameter-Efficient Prompting (EPE-P), which includes a comprehensive prompt matrix integrating information for all potential missing cases and developed specific prompt weight matrices for each modality to selectively extract relevant information from this comprehensive matrix. This approach not only compensates effectively for missing information but also reduces parameter redundancy. Extensive experiments demonstrate the effectiveness of our method. 

%In the future, we plan to investigate EPE-P with different multimodal architectures.

\noindent \textbf{Acknowledgement.} This work was supported by National Natural Science Foundation of China (Grant No. 62306061), and Guangdong Basic and Applied Basic Research Foundation (Grant No. 2023A1515140037). 

%\textbf{Future work: }We aim to extend the EPE-P approach to various backbone models beyond those explored in this study. Specifically, we plan to investigate the applicability of EPE-P with different multimodal learning architectures. Our goal is to develop a unified framework for evidence-based parameter-efficient prompting that can be seamlessly integrated into diverse models for handling missing modalities.

\clearpage
\bibliographystyle{IEEEbib}
\bibliography{strings,refs}

\end{document}